\documentclass[runningheads]{llncs}
\usepackage{graphicx}
\usepackage{float}
\usepackage[T1]{fontenc}
\usepackage[scaled=0.9]{beramono}
\usepackage{listings}
\usepackage{microtype}
\usepackage[table]{xcolor}
\usepackage{booktabs}
\usepackage{amssymb}
\usepackage{comment}

\newcommand*\LSTfont{\footnotesize\ttfamily}
\newcommand{\cmark}{\parbox[t]{2cm}{\centering \color{green}$\checkmark$}}
\newcommand{\xmark}{\parbox[t]{2cm}{\centering \color{red}$\times$}}

\begin{document}
\title{Enhancing Software-Related Information Extraction via Single-Choice Question Answering with Large Language Models}
%\title{Enhancing Software Related Information Extraction with Generative Language Models through Single-Choice Question Answering}
%"Single-Choice Question Answering with Large Language Models: A New Approach to Software Information Extraction"

\author{Wolfgang Otto\inst{1}\orcidID{0000-0002-9530-3631}
 \and
   Sharmila Upadhyaya\inst{1}\orcidID{0009-0003-7142-3887}
 \and
  Stefan~Dietze\inst{1,2}\orcidID{0009-0001-4364-9243} 
}  

\institute{
  GESIS -- Leibniz Institute for the Social Sciences, Cologne, Germany
    \email{\{firstname.lastname\}@gesis.org}
    \and Heinrich-Heine-University D\"usseldorf, Germany \\
    \email{stefan.dietze@hhu.de}
}

\authorrunning{Otto et al.}
\titlerunning{Enhancing Software IE via Single-Choice QA with LLMs}
\maketitle 

\begin{abstract}
This paper describes our participation in the Shared Task on Software Mentions Disambiguation~(SOMD), with a focus on improving relation extraction in scholarly texts through generative Large Language Models~(LLMs) using single-choice question-answering. The methodology prioritises the use of in-context learning capabilities of LLMs to extract software-related entities and their descriptive attributes, such as distributive information.  Our approach uses Retrieval-Augmented Generation~(RAG) techniques and LLMs for Named Entity Recognition~(NER) and Attributive NER to identify relationships between extracted software entities, providing a structured solution for analysing software citations in academic literature. The paper provides a detailed description of our approach, demonstrating how using LLMs in a single-choice QA paradigm can greatly enhance IE methodologies.
Our participation in the SOMD shared task highlights the importance of precise software citation practices and showcases our system's ability to overcome the challenges of disambiguating and extracting relationships between software mentions. This sets the groundwork for future research and development in this field.
\end{abstract}

\section{Introduction}
The evolution of information extraction (IE) in scholarly communications necessitates the development of innovative methodologies to accurately identify and categorise software mentions. This is a critical component for ensuring research transparency and reproducibility. The Shared Task on Software Mentions Disambiguation (SOMD) highlights the importance of refined citation practices amidst inconsistent referencing of software artifacts. This research employs generative Large Language Models (LLMs) to address the challenges in software mention extraction and relation identification, marking a significant step towards the sophisticated extraction of software-related information. Our approach integrates Retrieval-Augmented Generation (RAG) techniques with LLMs to dissect and comprehend the intricate web of software citations and their attributive details (e.g. the developer or the version) within scientific texts. By examining the potential of LLMs in performing IE tasks by transforming them into single-choice question-answering, we present a comprehensive analysis that addresses the nuances of software mention extraction and could be applied on a broader scope of scholarly artifacts NER and relation extraction. This paper explores the complexities of applying LLMs to NER tasks, providing insights into the challenges and proposing a new methodology for relation extraction that could pave the way for future innovations in the field.

\section{Related Work}
This section explores works in Named Entity Recognition (NER) and Information Extraction (IE), focusing on software mention extraction, scholarly artifact NER, and the use of generative LLMs in these areas.

In software mention extraction, the Softcite dataset~\cite{Du2021SoftciteDA} and the SoMeSci knowledge graph~\cite{Schindler2021SoMeSciA5} are significant. Softcite offers a gold-standard dataset for extracting software mentions from biomedical and economic research, while SoMeSci Knowledge Graph includes software mentions in scientific articles with relation labels such as version, developer and citations, highlighting the need for accurate software mention extraction.

In scholarly artifact NER, Saji and Matsubara~\cite{Saji2022ExtractingIA} introduced a method using academic knowledge graphs to extract research resource metadata from scholarly papers, enhancing metadata quality and repository size. Otto et al.~\cite{otto-etal-2023-gsap} developed the GSAP-NER corpus for extracting machine learning-related entities from scientific publications, filling the gap in general-purpose NER models. These words demostrate the importance of domain-specific NER tasks and leverageing knowledge graphs for enhancing research resource repositories.

Regarding LLMs for NER, Wang~\cite{wang2023gptner} explores the use of text-generation models for sequence labeling and underlying challenges, particularly in low-resource and few-shot setups. Furthermore, Xie~\cite{xie2024selfimproving} investigates LLMs' self-improving capabilities for zero-shot NER.

For LLMs in relation extraction, Wadhwa~\cite{wadhwa2023revisiting} uses few-shot prompting and fine-tuning with large language models achieving state-of-the-art performances in relation extraction using LLMs, while Wan~\cite{wan2023gptre} introduce GPT-RE to enhancing relation extraction accuracy through task-specific entity representations. 

Lastly, Xu~\cite{Xu2023LargeLM} conducts a comprehensive survey providing an overview of generative information extraction using LLMs, categorizing works by IE subtasks and learning paradigms, and highlighting the transformative potential of LLMs in IE.

Overall, these studies underscore the evolving methods and significant contributions in extracting information from scholarly texts using specialized datasets, domain-specific approaches, and advanced generative models.

\section{SOMD Shared Task}
The Shared Task on Software Mentions Disambiguation (SOMD) aims to enhance transparency and reproducibility in scientific research by improving software citation practices. Participants are required to develop models that can identify and disambiguate software mentions in scholarly texts using the expanded Software Mentions in Science (SoMeSci) knowledge graph, with a focus on AI and Computer Science. 

\textbf{Subtask 1}, Software NER, involves identifying four types of software-related entities: Application, Plugin, Operating System, and Programming Environment. Accurately classifying software mentions is necessary to understand their role in academic discourse.

\textbf{Subtask 2}, Attribute NER, aims to extract ten different attributive information associated with software entities, such as alternative names, abbreviations, authorship (developers), software release maintenance (versions, extensions, release dates, licenses), and elements of the referencing system (in-text citations, URLs, and software coreferences for cross-sentence linkage), to a specific software mention.
 
\textbf{Subtask 3}, Relation Extraction, involves establishing relationships between software entities and their attributes using specific relation types and mapping each attribute to these relations. This task aids in comprehending the interrelations of software entities within scholarly texts.

The tasks \textbf{evaluation metric}, is the Weighted Average Macro F1 score. This metric adjusts the influence of each label on the final result based on the number of test instances assigned to it, ensuring that labels with fewer instances have a proportionally smaller impact on the overall performance evaluation.

\section{Using LLMs for Software Related IE-Tasks}
The exploration of generative LLMs such as GPT-4 for Information Extraction (IE) tasks related to software underscores the transformative potential these models hold as general-purpose task solvers. The allure of leveraging LLMs in this capacity is significant, given their ability to process and generate human-like text across a wide range of topics and formats. However, the challenge arises in the specificity and nuanced requirements of domain-specific tasks, such as NER within specialized fields. It has been observed that, despite their vast knowledge and versatility, current models like GPT-4 often fall short when tasked with domain-specific NER, primarily due to their generalized training and lack of domain-specific tuning.

To mitigate these shortcomings and enhance the performance of LLMs in specialized IE tasks, various in-domain learning strategies are employed. These strategies are designed to equip the LLM with a deeper understanding of the task at hand, essentially guiding the model towards more accurate identification and classification of relevant text spans. Among these strategies, optimizing the task description plays a crucial role. A well-crafted, precise task description can significantly improve the model's focus and comprehension of the task's objectives, leading to more relevant and accurate outcomes.

Furthermore, the provision of speaking, prototypical examples serves as another effective strategy. By presenting the model with clear, illustrative examples that encapsulate the essence of the task, we can anchor its understanding and enhance its ability to generalize from these examples to new, unseen instances. This approach leverages the model's inherent learning capabilities, allowing it to draw parallels and apply learned concepts to the task at hand.

Additionally, augmenting the model's capabilities with Retrieval Augmented Generation (RAG) introduces a powerful dimension to the information extraction process. RAG combines the generative prowess of LLMs with the specificity and relevance of retrieved documents, enabling the model to access a broader context and detailed examples that are pertinent to the task. This strategy is particularly advantageous in domain-specific applications, where the relevance and accuracy of the information extracted are paramount.

In our approach, we capitalize on the last strategy, RAG, to maximize the utility of training sets for each specific task. The retrieval component of this strategy entails identifying instances within the training set that are similar to the test instance and can provide valuable insights for the identification and classification of relevant information. This method not only enhances the model's performance by providing it with task-relevant data but also ensures that the information extracted is of high relevance and quality, tailored to the specific demands of the domain-specific NER task. Through these tailored strategies, we aim to bridge the gap between the broad capabilities of LLMs and the precise requirements of domain-specific information extraction tasks, paving the way for more effective and efficient utilization of generative language models in specialized domains.

\subsection{Challenges in Applying LLMs to NER Tasks}
The integration of generative LLMs into NER~tasks introduces a set of unique challenges that can significantly impact the performance and reliability of extraction outcomes. One of the most prevalent issues in generative NER approaches is the phenomenon of hallucination\cite{wang2023gptner}, where the model generates entities not present in the test instance. This can result from the model misinterpreting the provided examples as part of the text from which entities should be extracted, leading to inaccuracies and inconsistencies in the results.
n
Furthermore, matching problems during the location of mention positions present considerable challenges, particularly in the context of span-based evaluation systems. These systems evaluate the accuracy of entity extraction based on the precise span of text identified as an entity. Discrepancies in the extracted span—whether through corrected spellings or variations in representation—can complicate the matching process. For example, the extraction of ``jquery'' as ``jQuery'' illustrates a common scenario where the prevalent spelling of a software library may differ from its mention in the text, yet both are correct. This variability necessitates sophisticated matching strategies to ensure accuracy in evaluation.

The situation is further complicated by texts containing multiple mentions of a single entity. The challenge lies in determining whether each mention can be accurately classified as an entity, alongside dealing with the ambiguity of different entities that are written in the same way. Overlapping or nested entity mentions that do not align with the ground truth data introduce additional layers of complexity, requiring nuanced approaches to entity recognition and classification.
Our proposed baseline solution, filtering out non-matching entities and employing rule-based decisions for handling multiple matches have been adopted. However, the disadvantage of this method is that it is based on simplistic heuristics.

Future work may explore advanced solutions like using LLMs for precise entity matching, offering potential improvements for NER challenges. While this paper does not delve into these complex methods, it highlights the importance of ongoing research to further refine and improve LLMs for more accurate and efficient entity extraction.

\subsection{Sample Retrieval for RAG on various IE-Tasks}

Retrieval-Augmented Generation (RAG) significantly bolsters the capabilities of LLMs in IE tasks by effectively utilizing both unstructured and structured contexts~\cite{gao2024retrievalaugmented,lewis2020retrieval}. This dual approach is essential in IE for achieving precise extractions, yet the selection of optimal samples for the generative process poses a substantial challenge. Addressing this involves two primary strategies: utilizing sentence embeddings to find contextually similar textual content for the LLM, and identifying analogous entities to uncover beneficial training sentences. These methods enable the LLM to discern structural and semantic patterns for more accurate text extractions. A crucial obstacle is accurately identifying target entities within test instances, for which we employ a pre-trained Language Model~(PLM) tailored to our NER task. This PLM is instrumental in both spotting potential entity candidates and facilitating entity similarity searches, leveraging last hidden state embeddings from training examples to locate matching entities within the dataset.

Our methodology extends to evaluating various retrieval techniques and their impact on the LLM’s efficiency, particularly within a Few-Shot learning framework. We explore different methods, including random illustrative samples, text similarity-based RAG, and entity-based sentence retrieval, to provide the LLM with contextually relevant examples, thereby optimizing the software entity extraction process from scientific texts. This exploration aims to identify the most effective strategies for utilizing LLMs in domain-specific software entity extraction tasks. By overcoming challenges related to resource demand, execution time, and precision in entity identification, our approach aims to enhance the accuracy and efficiency of NER processes, contributing valuable insights and methodologies to the field of computational linguistics and information extraction.

A special retrieval method is used for the RE Task. Because for this task, the entities for each test instance are given, we easily could list all possible relations (compare Table~\ref{tab:domain-range}). We could use them to find similar relations in the train set. If we find more then one, we decide for the one, with highest sentence similarity.

For the RE Task, a special retrieval method is employed. As entities for each test instance are provided, all possible relations can be listed based on Table~\ref{tab:domain-range}. These relations can then be used to find relations of the same type, and with the same domain and range in the train set. If multiple similar relations are found, the one with the highest sentence similarity is selected.
\subsection{Extraction of Software Entities}
The extraction of software entities from scientific texts represents a specialized challenge within the realm of NER, targeting the identification of software entities across four distinct categories ranging from applications to operating systems. Furthermore, this task extends beyond mere identification, seeking to understand the intent behind each mention of software entities—whether it pertains to creation, usage, deposition, or mere citation. 

We address this challenging task using LLMs, despite their high demands on resources and time, especially when processing extensive publications with sparse relevant text. Our approach includes a pipeline strategy that prioritizes selecting relevant text passages for LLM analysis, improving efficiency by filtering out unrelated content. This method's success depends on the selection accuracy, directly impacting recall. However, the trade-off for reduced computational costs justifies the potential minor decrease in recall.
Our performance optimization employs a hybrid method, combining a fine-tuned NER model for sentence selection with LLMs for information extraction. This approach faces limits, notably when sentences crucial for analysis are missed in the selection phase, capping the LLM extraction phase's accuracy as indicated by an initial sentence classification task recall of 0.882 (0.884 F1). This establishes a theoretical limit on extraction accuracy due to potential false negatives, illustrating a balance between efficiency and the precision constraints of LLMs in detailed text analysis.

\subsection{Extraction of Software Attributes}
Following the identification of software entities within scientific texts, a further nuanced aspect of NER and IE tasks emerges in the extraction of associated software attributes. These attributes encompass a wide array of specific details, the version, developer, citations, URLs, release dates, abbreviations, licenses, extensions, software co-references, and alternative names. 
We used a similar approach as for subtask 1, and utilised train sample retrieval to augment the task description in a few shot setup.
For each sample, including those derived from few-shot learning, the process entails presenting the sentence containing the software entity(ies) and then predicting a JSON list of identified entities along with their respective attribute types.

\subsection{Relation Extraction as Single-Choice Question Answering Task}
In the field of Natural Language Processing (NLP), the extraction of relations between entities within a text corpus poses significant challenges.
This study proposes a novel approach by conceptualizing the task of relations extraction as a single-choice question-answering (QA) activity. 
This method entails generating a comprehensive list of all possible entities within a sentence, drawing from the existing entities and their relationships as delineated in the training dataset. 
Each potential pair of entities is then evaluated to ascertain if a specific relation, such as ``version\_of'', appropriately links them. 
For instance, considering the relationship ``version\_of'', a sentence may be formulated as ``8 is the version of SPSS'', representing a possible relation between the version number and the software entity.

For every sentence in the dataset, this process yields a set of single-choice questions, each positing a potential relationship between entities. 
These questions are then prompted to a LLM for answering. 
The LLM's task is to select the most plausible relation from among the given options, thereby facilitating the extraction of accurate entity relations from the text. 
However, this approach is not without its challenges. 
A primary source of error stems from instances where multiple relations could plausibly link a pair of entities, leading to ambiguity and complicating the single-choice question-answering framework.
Despite the challenges, we demonstrate that treating relation extraction as a single-choice QA task provides a structured and innovative approach to extracting valuable insights from complex textual data.

\section{Experiments}
\subsection{Models} 
\textbf{Fine-tuned Model} In our experimental setup, the initial phase focused on fine-tuning a language model specifically for Subtask 1, which involved NER. Following the methodology similar to Schindler \cite{Schindler2021SoMeSciA5}, we employed the SciBERT~model~\cite{Beltagy2019SciBERTAP}, given its pre-training on scientific corpus, making it apt for NER fine-tuning within scholarly texts. The fine-tuning process entailed rigorous parameter optimization, including adjustments to the batch size, learning rate, and the relative share of negative samples—sentences that do not contain any annotations. This optimization utilized a 90/10 train/evaluation data split from the available training dataset. Subsequent to parameter tuning, we conducted a final training run with a modified data split of 95/5 train/evaluation to maximize the training data's utility. \\
\textbf{Generative Large Language Models}
For a comparative analysis, our study integrated LLMs, specifically examining the performance differences between GPT-3.5-fast and GPT-4-fast models accessed via the OpenAI API. To ensure deterministic outputs for comparison, we set the temperature parameter to zero, eliminating randomness in the model's response generation.

\subsection{Prompting}

\textbf{Software NER} The prompting strategy for Software NER involved providing a concise task description, which included specifying the task as NER and intention classification, alongside an enumeration of target labels. We also highlighted the domain specificity of the texts (i.e., scientific publications) and requested the output in JSON format, delineating separate labels for entity type and intention. Sample sentences from the training set, along with their corresponding JSON output, were included as illustrative examples. This setup varied in the number and order of displayed examples based on the retrieval method, exploring the impact of these factors on model performance through separate experiments.

\textbf{Attributive NER} For Attributive NER, the prompt construction mirrored the approach taken in Subtask 1 but incorporated more detailed rules, emphasizing the necessity for attributes to relate directly to software entities. Known software entities were additionally provided as input, even for test instances, adhering to the oracle setup defined in the shared task guidelines. This method aimed to refine the model's ability to extract and classify attributive information accurately.

\textbf{Relation Extraction} The approach to Relation Extraction reimagined the task as a series of single-choice question-answering challenges. Each potential relation, given the software and attributive entities within a sentence, was listed, with claims formulated for each possible relationship (e.g., "IBM is the developer of Windows"). The model was tasked with identifying the veracity of each claim through a single-choice question format, where all claims were enumerated and solutions provided in a batch format for each example sentence. This setup culminated in presenting the test instance alongside its single-choice questions, expecting the model to deliver decisions on the relational claims.

\subsection{Train Sample Retrieval for Few-Shot Generation}
Our experimental framework explored various methods for test sample retrieval to ascertain the most effective approach in enhancing model performance. These methods included the use of random illustrative samples to represent every possible signature, retrieval based on entity similarity, and retrieval based on sentence similarity. Each method was evaluated against a baseline to determine its impact on the accuracy and efficiency of the information extraction tasks at hand.

\subsection{Relation Extraction Baseline}
In the development of our baseline for relation extraction, we established a robust heuristic framework derived from an analysis of potential relations indicated within the text. Our strategy was guided by two principal rules aimed at simplifying the decision-making process for identifying accurate relationships between entities. Firstly, we limited our consideration to relations that necessitate the presence of at least one related entity, deliberately excluding optional inter-software entity relations such as "specification\_of" and "PlugIn\_of." Given the infrequent occurrence of these cases within the training dataset, we anticipated only a minimal impact on the overall performance metric, specifically the weighted mean macro F1 score for the subtask. Secondly, for all remaining relation types, our approach favored selecting the closest possible entity positioned to the left of the focal software entity as the most likely relation partner. This heuristic was not only straightforward but proved to be highly effective, aligning our baseline performance with that of the top contenders in the shared task. This methodology highlights the potential of leveraging simple, rule-based strategies to achieve competitive results in complex relation extraction challenges.

\section{Results} 

\begin{table}[t]
\centering
\caption{Results of Different Models and Retrieval Methods on Subtask 1}
\label{tab:results}
\begin{tabular}{lllll}
\hline
\textbf{Paradigm} & \textbf{Retrieval} & \textbf{F1} & \textbf{Model} & \textbf{parameter} \\
\hline
Finetuned & - & 0.599 & SciBERT & - \\
\hline
Prompt & Random & 0.483 & GPT 3-5 & Random k=7 \\
Prompt & Random & 0.525 & GPT 3-5 & Random all entity types shown \\
Prompt & Sim. sentences & 0.647 & GPT 3.5 & topn=10 \\ % (mean)
Prompt & Sim. entities  & 0.624 & GPT 3.5 & topn=7 x n entities \\
\hline
Prompt & Random & 0.574 & GPT 4 & Random all entity types shown \\
Prompt & Sim. sentences & 0.677 & GPT 4 & topn=10 \\ %  (tfidf) 
Prompt & Sim. entities  & \textbf{0.679} & GPT 4 & topn=7 x n entities\\
\hline
\end{tabular}
\end{table}
Our analysis in Subtask 1 (Software NER) shows a varied performance landscape across different models and retrieval methods (Table~\ref{tab:results}). The finetuned baseline, which uses SciBERT, achieved a solid foundation with a 59.9\% F1 score. However, LLMs that use random samples without fine-tuning showed a decrease in performance, with the highest F1 score reaching only 57.4\%.\\
A closer examination of retrieval-based models indicates that LLMs perform better. The highest F1 score of 67.9\% was achieved by sentence similarity retrieval
models, while entity retrieval showed the best performance at 67.7\% F1.
The transition from GPT 3.5 to GPT 4 models resulted in a significant improvement of approximately 3--5\%, although it required around three times more computation time. Notably, our best models were able to perform within a mere 3\% below the theoretical maximum by utilizing SciBERT for sentence selection, leveraging oracle positive sentences.\\
In Subtask 2 (Attributive NER), our methodology showed a significant improvement of +10\% in F1 performance compared to our competitors, demonstrating the effectiveness of our approach in a low data regime (Table~\ref{tab:somd_results}). For Subtask 3 (Relation Extraction), our LLM Single-Question Answering model further improved the F1 score by 5.1\%, building on the already competent performance of the heuristic baseline and highlighting the advantage of our method. Furthermore, it has been demonstrated that using 7-10 samples is the most effective strategy, as it optimises the balance between input complexity and model performance.\\
This analysis highlights the potential of utilising advanced LLM techniques and carefully selected retrieval methods to significantly improve the accuracy and efficiency of NER tasks in specialised domains.

\begin{table}[tb]
\caption{SOMD Performance Rankings (Weighted Average Macro)}
\label{tab:somd_results}
\begin{tabular}{@{}cclccc|ccclcccc@{}}
\toprule
Task & \# & User   & F1 & Precision  & Recall & Task & \# & User   & F1 & Precision  & Recall \\
\midrule
1 & 1    & \textbf{phinx}            & \textbf{.740}     & \textbf{.761}            & \textbf{.750}        & 2 & 1    & \textbf{\emph{ours}}  & \textbf{\emph{.838}}    & \textbf{\emph{.835}}             & \textbf{\emph{.847}}        \\
1 & 2    & david-s477       & .692      & .739            & .711          & 2 & 2    & phinx        & .743      & .745             & .748          \\
1 & 3    & ThuyNT03         & .678      & .729           & .649         & 3 & 1    & \textbf{\emph{ours}}       & \textbf{\emph{.916}}      & \emph{.911}            & \textbf{\emph{.924}}          \\
1 & 4    & \emph{ours}    & \emph{.652}     & \emph{.679}           & \emph{.664}  &    3 &2    & phinx                &  .897   & .900            & .897          \\ 
1 & 5    & vampire           & .648      & .682            & .637      &   3 &-    & \emph{baseline}        &    \emph{.864}  & \emph{.857}   & \emph{.875}      \\ 
& & & & & & 3 & -   & \emph{necessary}          & \emph{.562}  & \textbf{\emph{.933}}      & \emph{.415}   \\
\bottomrule
\end{tabular}
\end{table}

\section{Conclusion}
Our research on enhancing Relation Extraction (RE) with LLMs through Single-Choice QA has introduced a novel intersection of methodologies aimed at improving the precision of information extraction in the context of scientific texts. By integrating Retrieval-Augmented Generation (RAG) with LLMs and adopting a methodical approach to fine-tuning and leveraging large language models such as SciBERT and use the model to support GPT variants, we have demonstrated the capability of LLMs to navigate the complexities inherent in the extraction of software entities and their attributes.

The exploration of different retrieval strategies—ranging using entity and sentence similarity underscores our commitment to refining the inputs for generative models, ensuring that they are fed the most relevant and contextually appropriate data. This meticulous preparation has allowed us to significantly boost the performance of LLMs in recognising nuanced distinctions among software-related entities and accurately extracting relation types within scholarly articles. Our experiments have not only highlighted the efficacy of LLMs in addressing domain-specific tasks with a limited set of examples but also revealed the inherent challenges, such as the difficulties in matching entity mentions accurately. Despite these hurdles, our single-choice QA approach for RE lead to a strong heuristic baseline for relation extraction and show how altering the task simplifies the problem.

The outcomes of our research indicate a promising direction for future work in leveraging LLMs for NER and RE tasks. The development of our system for participation in the SOMD shared task has illustrated the potential of a single-choice QA approach to relation extraction, offering a structured and scalable method for extracting meaningful insights from textual data. Our findings contribute to the growing body of knowledge on the application of generative models in the field of computational linguistics, paving the way for more sophisticated and efficient methodologies in information extraction from scientific literature.

\section*{Acknowledgements}
We thank the anonymous reviewers for their constructive feedback.
This work has been partially funded by the Deutsche Forschungsgemeinschaft (DFG, German Research Foundation) as part of the Projects BERD@NFDI (grant~number~460037581) as well as NFDI4DS (grant~number~460234259).

\newpage
\appendix

\section{Dataset Overview}
Each subtask in the SOMD Shared Task is evaluated independently.
Subtask 2’s attributive NER uses the identified software entities from Subtask 1 as
inputs, and the annotations from the first two subtasks are combined for Subtask 3’s Relation Extraction. The dataset is hierarchically structured, with each subtask’s test set forming a subset of the previous one. This lead to data leakage in Subtask 1 and to a lesser extent in Subtask 2. This design ensures that tasks are interconnected, but it requires careful interpretation of results, particularly with respect to error propagation.
Although this design creates data leakage, it ensures the interconnectedness and cumulative nature of the challenges.
The initial presentation of the Corpus is as the SoMeSci knowledge graph \cite{Schindler2021SoMeSciA5}.
It is a comprehensive dataset consisting of 1,367 documents and containing 399,942 triples that represent 47,524 sentences. The dataset includes 2,728 software entities, totaling 7,237 labeled entities.
The dataset includes positive and negative examples, i.e. sentences without software mentions, which improves the accuracy of the model. The presence of duplicate sentences, headings, and varying sentence
lengths increases the complexity of the text and makes extraction tasks from scientific texts more challenging.
For possible domains and ranges of the relations int the train set compare Table~\ref{tab:domain-range}
\begin{table}[t]
\centering
\caption{Object Properties, Domain, and Range Classes with Inner Software Relations.}
\label{tab:domain-range}
\begin{tabular}{@{}lccccc@{}}
\toprule

\textbf{Range Class} & \textbf{Application} & \textbf{Operating} & \textbf{PlugIn} & \textbf{Programming} \\ 
\textbf{$\Rightarrow$ Object Property} & & \textbf{System} &  & \textbf{Environment} \\ 
\midrule
\parbox[t]{3.5cm}{Abbreviation \\ $\Rightarrow$ Abbreviation\_of}  & \cmark & \cmark & \cmark & \cmark \\
\parbox[t]{3.5cm}{Developer \\ $\Rightarrow$ Developer\_of } & \cmark & \cmark & \cmark & \cmark\\
\parbox[t]{3.5cm}{Release \\ $\Rightarrow$ Release\_of}  & \cmark & \cmark & \cmark & \cmark \\
\parbox[t]{3.5cm}{Version \\ $\Rightarrow$ Version\_of} & \cmark & \cmark & \cmark & \cmark \\
\parbox[t]{3.5cm}{Citation \\ $\Rightarrow$ Citation\_of} & \cmark & \xmark & \cmark & \cmark  \\
\parbox[t]{3.5cm}{PlugIn \\ $\Rightarrow$ PlugIn\_of} & \cmark & \xmark & \cmark & \cmark \\
\parbox[t]{3.5cm}{URL \\ $\Rightarrow$ URL\_of} & \cmark & \xmark & \cmark & \cmark  \\
\parbox[t]{3.5cm}{Extension \\ $\Rightarrow$ Extension\_of} & \cmark & \cmark & \xmark & \xmark \\
\parbox[t]{3.5cm}{License \\ $\Rightarrow$ License\_of} & \cmark & \xmark & \cmark & \xmark \\
\parbox[t]{3.5cm}{AlternativeName \\ $\Rightarrow$ AlternativeName\_of} & \cmark & \xmark & \xmark & \xmark  \\
\midrule
\addlinespace % Adds some space before the inner software relations for clarity
\parbox[t]{3.5cm}{Application \\ $\Rightarrow$ PlugIn\_of \\ $\Rightarrow$ Specification\_of} & \cmark & \xmark & \xmark & \xmark  \\
\parbox[t]{3.7cm}{ProgrammingEnvironment \\ $\Rightarrow$ Specification\_of }& \xmark & \xmark & \xmark & \cmark  \\
\bottomrule
\end{tabular}
\end{table}

\section{Similarity Search Examples}
\subsection{Search By Entity Similarity}

\begin{table}[H]
\caption{Retrieval example based on entity similarity for RAG. The first row is the query entity. ``sim'' reflects the cosine similarity of the enities. }
\label{table:sim_ent}
\begin{tabular}{lllllll}
\toprule
        entity &        label &  sim & split &  sentence \\
\midrule
   PhosphOrtholog &  Application  &    1.00 &  test &
             To this end , we have developed an \\
 & Creation & & &
             automated web - based tool ,\\

 & & & &  PhosphOrtholog , which allows batch \\
 & & & &  processing and mapping of large species -\\
 & & & &  specific PTM datasets to compare overlap\\
 & & & &  at a site - specific level .\\
         SNPdetector &  Application &    0.93 &  train & 
            We developed a software tool , \\
&  Creation & & & 
            SNPdetector , for automated\\
 & & & &  identification of SNPs and mutations in \\
 & & & &  fluorescence - based resequencing reads . \\
  ESPRIT - Forest &  Application  &    0.92 &   train  & 
            In this paper we developed a new\\
 & Creation  & & &
            algorithm called ESPRIT - Forest for\\
 & & & &  parallel hierarchical clustering of sequences . \\
\bottomrule
\end{tabular}
\end{table}

\subsection{Search By Sentence Similarity}
\begin{table}[H]
\caption{Retrieval example based on sentence similarity for RAG. The first row is the query entity. ``sim'' reflects the cosine similarity of the sentences. No information about the annotated entities are given for retrieval. The example sentence is the same as in Table~\ref{table:sim_ent}}
\label{table:sim_sent}
\begin{tabular}{cll}
\toprule
split &  sim &            text \\
\midrule
 test &    1.00 &
     To this end , we have developed an automated web - based tool , \\
 & & PhosphOrtholog , which allows batch processing and mapping of large \\
 & & species - specific PTM datasets to compare overlap at a site - \\
 & & specific level . \\
train &    0.99 &                                                                                      Here 
     we present a tool , Podbat ( Positioning database and analysis\\
 & & tool ) , that incorporates data from various sources and allows\\
 & & detailed dissection of the entire range of chromatin modifications\\
 & & simultaneously . \\
train &    0.99 & 
     We designed and developed a new method , MSACompro , to\\
 & & synergistically incorporate predicted secondary structure ,\\
 & & relative solvent accessibility , and residue - residue contact \\
 & & information into the currently most accurate posterior probability \\
& & - based MSA methods to improve the accuracy of mult... \\
train &    0.99 &   
     In this paper , we present a lossless compression tool , MAFCO ,\\
 & & specifically designed to compress MAF ( Multiple Alignment Format )\\
 & & files . \\
\bottomrule
\end{tabular}
\end{table}

\section{Prompting Examples}
\begin{figure}[H]
\caption{Software NER Few-Shot prompt (n=2). The shown sample is the same as in the similarity examples in Table~\ref{table:sim_ent} and~\ref{table:sim_sent}.}
\begin{lstlisting}[label=lst:prompt_ner]
# Task: Exctract entities in a closed NER setting and classify their intention.
## Entity Types (closed setting):
The entity types in this closed setting are: ['Application', 'ProgrammingEnvironment', 'SoftwarePackageOrPlugin', 'OperatingSystem', 'SoftwareCoreference']
Do not include version information for the entities.
if no entity of the given entity types in the context return an empty list

## Intention Classification:
For each entity mention classify the intention of the usage in the context.
Use only and exact one intention per entity.
The intention classes are: ['Usage', 'Deposition', 'Creation', 'Mention']

## The Contexts:
The contexts to extract the entities are about software and from scientific publications.

## Output Format:
Return the result in a json list of objects.

# Examples:
## Context:
"""We developed a software tool , SNPdetector , for automated identification of SNPs and mutations in fluorescence - based resequencing reads ."""
## Detected Named Entities with Intention:
[   {
		"text": "SNPdetector",
		"label": "Application",
		"intention": "Creation" } ]
## Context:
"""In this paper we developed a new algorithm called ESPRIT - Forest for parallel hierarchical clustering of sequences ."""
## Detected Named Entities with Intention:
[   {
		"text": "ESPRIT - Forest",
		"label": "Application",
		"intention": "Creation" } ]
## Context:
"""To this end , we have developed an automated web - based tool , PhosphOrtholog , which allows batch processing and mapping of large species - specific PTM datasets to compare overlap at a site - specific level ."""
## Detected Named Entities with Intention:
\end{lstlisting}
\end{figure}

\begin{figure}[H]
\caption{Attibutive NER Few-Shot prompt (n=2).}
\begin{lstlisting}[label=lst:prompt_attributive_ner]
# Task: 
## Task Description
Exctract attributive text spans about software related entities in a closed attibutive NER setting.

## The Contexts:
The contexts to extract the attributive text spans are about software and from scientific publications.

## The Software Related Entities:
The software related entities are given as a list of json object with mention and entity type.
Do not annotate any related software entities.

## Attribute Types (closed setting):
The attribute types to identify in this closed setting are:
 ['Version', 'Developer', 'Citation', 'URL', 'Release', 'Abbreviation', 'License', 'Extension', 'AlternativeName']
Keep the "Version" note separate, without overlap with other entities or attribute spans.
No span overlaps are allowed inbetween attributive spans of to the given software related entites.
Note that the attribute types must be associated with a software-related entity of one of the following types:  ['Application', 'ProgrammingEnvironment', 'PlugIn', 'OperatingSystem', 'SoftwareCoreference'].
To help you, you will receive all software-related entities as json for each context.

## Output Format:
Return the result in a json list of objects.

# Examples:
## Context:
"""LAMINA uses the Java Advanced Imaging ( JAI ) package http://java.sun.com/javase/technologies/desktop/media/jai/ to support common image file formats , which is bundled with the installation and hence no additional installation should be required ."""
### Detected Software Related Named Entities:
[{
    "text": "LAMINA",
    "label": "Application"
 }, {
    "text": "Java Advanced Imaging",
    "label": "Application"}]

### Detected Attributive Text Spans:
[{
    "text": "JAI",
    "label": "Abbreviation"
 }, {
    "text": "http://java.sun.com/javase/technologies/desktop/media/jai/",
    "label": "URL"}]

\end{lstlisting}
\end{figure}

\begin{figure}[H]
\caption{Single-Choice QA prompt for Relation Annotation (n=2).}
\begin{lstlisting}[label=lst:prompt_single-choice_qa]
# Answer the following single-choice questions on sentences from scientific publications on the subject of software.
 * Please note that there is exactly one answer to each question.
## Sentence:
"""The web service [44] available at http://www.fragit.org enables users to upload their structure , fragment it and download the resulting input file to GAMESS ."""
### Question(s):
Which of the statements about the Citation '[44]' is true?
idx_0: '[44]' is the citation of 'web service'.
idx_1: '[44]' is the citation of 'GAMESS'.

Which of the statements about the URL 'http://www.fragit.org' is true?
idx_2: 'http://www.fragit.org' is the url of 'web service'.
idx_3: 'http://www.fragit.org' is the url of 'GAMESS'.

### Answers to the questions about the sentence(s):
idx_0: True
idx_1: False
idx_2: True
idx_3: False

## Sentence:
"""ASPASIA , released under the Artistic License ( 2.0 ) , can be downloaded from http://www.york.ac.uk/ycil/software ."""
### Question(s):
Which of the statements about the Version '2.0' is true?
idx_4: '2.0' is the version of 'ASPASIA'.
idx_5: '2.0' is the version of 'Artistic License'.

Which of the statements about the URL 'http://www.york.ac.uk/ycil/software' is true?
idx_6: 'http://www.york.ac.uk/ycil/software' is the url of 'ASPASIA'.
idx_7: 'http://www.york.ac.uk/ycil/software' is the url of 'Artistic License'.

### Answers to the questions about the sentence(s):
idx_4: False
idx_5: True
idx_6: True
idx_7: False

## Sentence:
"""The web - tool can be accessed from http://www.phosphortholog.com/ using any typical web browser ( except Internet Explorer ) ."""

### Question(s):
Which of the statements about the URL 'http://www.phosphortholog.com/' is true?
idx_8: 'http://www.phosphortholog.com/' is the url of 'web - tool'.
idx_9: 'http://www.phosphortholog.com/' is the url of 'Internet Explorer'.

### Answers to the questions about the sentence(s):
\end{lstlisting}
\end{figure}

\bibliographystyle{splncs04}
\bibliography{references}

\begin{thebibliography}{10}
\providecommand{\url}[1]{\texttt{#1}}
\providecommand{\urlprefix}{URL }
\providecommand{\doi}[1]{https://doi.org/#1}

\bibitem{Beltagy2019SciBERTAP}
Beltagy, I., Lo, K., Cohan, A.: Scibert: A pretrained language model for scientific text. In: Conference on Empirical Methods in Natural Language Processing (2019). \doi{10.18653/v1/d19-1371}

\bibitem{Du2021SoftciteDA}
Du, C.F., Cohoon, J., Lopez, P., Howison, J.: Softcite dataset: A dataset of software mentions in biomedical and economic research publications. Journal of the Association for Information Science and Technology  \textbf{72},  870 -- 884 (2021). \doi{10.1002/asi.24454}

\bibitem{gao2024retrievalaugmented}
Gao, Y., Xiong, Y., Gao, X., Jia, K., Pan, J., Bi, Y., Dai, Y., Sun, J., Wang, M., Wang, H.: Retrieval-augmented generation for large language models: A survey (2024). \doi{10.48550/arXiv.2312.10997}

\bibitem{lewis2020retrieval}
Lewis, P., Perez, E., Piktus, A., Petroni, F., Karpukhin, V., Goyal, N., K\"{u}ttler, H., Lewis, M., Yih, W.t., Rockt\"{a}schel, T., Riedel, S., Kiela, D.: Retrieval-augmented generation for knowledge-intensive nlp tasks. In: Larochelle, H., Ranzato, M., Hadsell, R., Balcan, M., Lin, H. (eds.) Advances in Neural Information Processing Systems. vol.~33, pp. 9459--9474. Curran Associates, Inc. (2020)

\bibitem{otto-etal-2023-gsap}
Otto, W., Zloch, M., Gan, L., Karmakar, S., Dietze, S.: {GSAP}-{NER}: A novel task, corpus, and baseline for scholarly entity extraction focused on machine learning models and datasets. In: Bouamor, H., Pino, J., Bali, K. (eds.) Findings of the Association for Computational Linguistics: EMNLP 2023. pp. 8166--8176. Association for Computational Linguistics, Singapore (Dec 2023). \doi{10.18653/v1/2023.findings-emnlp.548}, \url{https://aclanthology.org/2023.findings-emnlp.548}

\bibitem{Saji2022ExtractingIA}
Saji, A., Matsubara, S.: Extracting information about research resources from scholarly papers. In: International Conference on Asian Digital Libraries. pp. 440--480 (2022). \doi{10.1007/978-3-031-21756-2\_35}

\bibitem{Schindler2021SoMeSciA5}
Schindler, D., Bensmann, F., Dietze, S., Kr{\"u}ger, F.: Somesci- a 5 star open data gold standard knowledge graph of software mentions in scientific articles. Proceedings of the 30th ACM International Conference on Information \& Knowledge Management pp. 4574--4583 (2021). \doi{10.1145/3459637.3482017}

\bibitem{wadhwa2023revisiting}
Wadhwa, S., Amir, S., Wallace, B.: Revisiting relation extraction in the era of large language models. In: Rogers, A., Boyd-Graber, J., Okazaki, N. (eds.) Proceedings of the 61st Annual Meeting of the Association for Computational Linguistics (Volume 1: Long Papers). pp. 15566--15589. Association for Computational Linguistics, Toronto, Canada (Jul 2023). \doi{10.18653/v1/2023.acl-long.868}, \url{https://aclanthology.org/2023.acl-long.868}

\bibitem{wan2023gptre}
Wan, Z., Cheng, F., Mao, Z., Liu, Q., Song, H., Li, J., Kurohashi, S.: {GPT}-{RE}: In-context learning for relation extraction using large language models. In: Bouamor, H., Pino, J., Bali, K. (eds.) Proceedings of the 2023 Conference on Empirical Methods in Natural Language Processing. pp. 3534--3547. Association for Computational Linguistics, Singapore (Dec 2023). \doi{10.18653/v1/2023.emnlp-main.214}, \url{https://aclanthology.org/2023.emnlp-main.214}

\bibitem{wang2023gptner}
Wang, S., Sun, X., Li, X., Ouyang, R., Wu, F., Zhang, T., Li, J., Wang, G.: Gpt-ner: Named entity recognition via large language models (2023). \doi{10.48550/arXiv.2304.10428}

\bibitem{xie2024selfimproving}
Xie, T., Li, Q., Zhang, Y., Liu, Z., Wang, H.: Self-improving for zero-shot named entity recognition with large language models (2024). \doi{10.48550/arxiv.2311.08921}

\bibitem{Xu2023LargeLM}
Xu, D., Chen, W., Peng, W., Zhang, C., Xu, T., Zhao, X., Wu, X., Zheng, Y., Chen, E.: Large language models for generative information extraction: A survey. ArXiv  (2023). \doi{10.48550/arxiv.2312.17617}

\end{thebibliography}

\begin{comment}
## Context:
"""The library , called ENCORE ( http://encore-similarity.github.io/encore ) , interfaces with the MDAnalysis molecular analysis toolkit [14] and can be used both as a Python library and from the command line ."""

### Detected Software Related Named Entities:
[
	{
		"text": "ENCORE",
		"label": "PlugIn"
	},
	{
		"text": "MDAnalysis molecular analysis toolkit",
		"label": "PlugIn"
	},
	{
		"text": "Python",
		"label": "ProgrammingEnvironment"
	}
]

### Detected Attributive Text Spans:
[
	{
		"text": "http://encore-similarity.github.io/encore",
		"label": "URL"
	},
	{
		"text": "[14]",
		"label": "Citation"
	}
]

## Context:

"""A Virtual Machine for Oracle ' s VirtualBox has also been built to provide easy access to IDEPI for users unfamiliar with the intricacies of Python package management , and is available from the main package distribution page ( http://github.com/veg/idepi/ ) ."""

### Detected Software Related Named Entities:
[
	{
		"text": "VirtualBox",
		"label": "Application"
	},
	{
		"text": "IDEPI",
		"label": "PlugIn"
	},
	{
		"text": "Python",
		"label": "ProgrammingEnvironment"
	}
]

### Detected Attributive Text Spans:

\end{comment}

\end{document}